\documentclass[12pt]{article}

\usepackage[utf8]{inputenc}          
\usepackage[T1]{fontenc}             
\usepackage{times}                   
\usepackage{geometry}                
\usepackage{tikz}
\usepackage{setspace}                
\usepackage{graphicx}                
\usepackage{caption}                 
\usepackage{subcaption}              
\usepackage{booktabs}                
\usepackage{multirow}                
\usepackage{amsmath, amssymb}        
\usepackage{natbib}                  
\usepackage{hyperref}
\usepackage{xurl}                     
\usepackage{algorithm}
\usepackage{algorithmic}
\usepackage{longtable, array}        

\hypersetup{
    colorlinks=true,       
    linkcolor=blue,  
    citecolor=green!50!black, 
    urlcolor=blue 
}

\title{Introducing Flexible Monotone Multiple Choice Item Response Theory Models and Bit Scales}

\author{
    \textbf{Joakim Wallmark}\textsuperscript{1},
    \textbf{Maria Josefsson}\textsuperscript{1},
    \textbf{Marie Wiberg}\textsuperscript{1} \\
    \textsuperscript{1}Department of Statistics, USBE, Umeå University, Sweden
}

\date{}

\begin{document}
\maketitle

\begin{abstract}
    \noindent
    Item Response Theory (IRT) is a powerful statistical approach for evaluating test items and determining test taker abilities through response analysis. An IRT model that better fits the data leads to more accurate latent trait estimates. In this study, we present a new model for multiple choice data, the monotone multiple choice (MMC) model, which we fit using autoencoders. Using both simulated scenarios and real data from the Swedish Scholastic Aptitude Test, we demonstrate empirically that the MMC model outperforms the traditional nominal response IRT model in terms of fit. Furthermore, we illustrate how the latent trait scale from any fitted IRT model can be transformed into a ratio scale, aiding in score interpretation and making it easier to compare different types of IRT models. We refer to these new scales as bit scales. Bit scales are especially useful for models for which minimal or no assumptions are made for the latent trait scale distributions, such as for the autoencoder fitted models in this study.
\end{abstract}

\section{Introduction}\label{sec-intro}

Item Response Theory (IRT) models \citep{van2016a}, which model the probability of various item responses based on an underlying latent trait, are extensively used to address practical issues in testing. These problems range from scoring test takers, equating test forms and determining the psychometric characteristics of test forms and items, to enhancing test delivery efficiency in more intricate assessment systems. 

The fundamental assumption of IRT is that the probability for each response to each item in a test or questionnaire is a function of the latent trait(s) being measured, the item response function (IRF). Both the latent traits and the IRFs are estimated from the observed item responses. In numerous IRT applications, there is a prevailing assumption that the latent trait being measured by the test follows a normal distribution \citep[see e.g.,][]{martin2016,huang2022}. However, as the latent trait scale is constructed when fitting the IRT model, it can always be re-scaled freely, provided that the ordering of different values on the scale remains unchanged. In fact, any monotonic transformation that retains the same ordering of the estimated latent trait scores can be applied, producing an equally valid model \citep[see e.g.,][]{lord1975, wallmark2023}. As a consequence, distances on the latent trait scale of a fitted IRT model hold no inherent meaning. Thus the typical assumption about a normally distributed latent trait serves mostly as a constraint to aid in the convergence of the model fitting algorithm, one of the more common ones being marginal maximum likelihood \citep[MML,][]{bock1981}. While aiding in convergence and interpretation, such constraints may also limit the ability for a chosen model to accurately fit the data.

Moreover, the IRFs are typically modeled using established functional forms. For example, a common modeling assumption is that the link between the latent trait and the likelihood of responding to dichotomous items can be represented by a logistic function, for instance, the two-parameter logistic (2PL) model \citep{birnbaum1968}. When addressing items with more than 2 possible responses, the IRFs are often extensions of the logistic function to accommodate multiple categories. Examples include the generalized partial credit model \citep{muraki1992} or the nominal response (NR) model \citep{bock1972}. Usually the assumptions about a normally distributed latent trait together with the standard parametric forms of the IRFs do not fully hold. In such situations, adhering to standard modeling techniques leads to inaccurate retrieval of the true IRFs and latent trait estimates \citep{falk2016, wiberg2019, wallmark2023}. Various methodologies have been proposed to estimate IRT models when these assumptions do not hold. These include approaches for modeling either a non-normally distributed latent trait \citep{mislevy1984, woods2008, woods2009}, a non-standard IRF \citep{ramsay1991, falk2016} or both \citep{ramsay2020, wallmark2023}. In order to improve the interpretability of models making no specific assumptions on the latent trait distribution, \citet{ramsay2020} utilized concepts from information theory to define what they referred to as the scale curve. Test taker locations on the scale curve were then used as measures of the latent trait. However, a limitation with the scale curve is that it goes to infinity if the IRF for any item response tends towards zero. This makes it unsuitable for most models, such as the 2PL, NR or generalized partial credit models to name a few.

In this research, we propose a new model for modeling multiple choice test items which we refer to as the monotone multiple choice (MMC) model. Similar to the multiple choice model introduced by \citet{samejima1979}, the model not only models the probability for answering each item correctly, but also the probability for selecting each incorrect option (distractor) of an item over various levels of the latent trait. This provides a tool for  test creators to evaluate the quality of the distractors for each item. Furthermore, it allows for more accurate latent trait estimation, as the model can take the probability of selecting each distractor over different levels of the latent trait into account. 

To fit MMC models, we use autoencoders \citep[AEs,][]{kramer1991}. An AE is a type of neural network used for unsupervised learning, often aimed at dimensionality reduction or efficient data encoding. In an IRT context, AEs allow for estimating both non-normal latent trait distributions and non-parametric IRFs simultaneously. Most previous studies using AEs in an IRT context have focused specifically on variational autoencoders \citep[VAEs,][]{wu2020, wu2022, urban2021}, with the exception of \citet{paaßen2022} who used non-variational AEs to fit multidimensional parametric models. VAEs place a penalty on the loss function forcing the latent variable distribution closer to a normal. This allows one to sample the latent distribution and generate new data points similar to old ones. While this is useful in other applications, such as image generation, it is not needed in IRT. In IRT, the primary focus is not to generate new test scores, but to get a well fitting model and score the test takers on an underlying latent variable. Consequently, this study utilizes non-variational AEs to fit IRT models, prioritizing model accuracy and more precise scoring of latent traits.

Additionally, we improve upon the scale curve introduced by \citet{ramsay2020} and introduce what we refer to as bit scales. Just like the scale curves, bit scales are ratio scales, with an absolute zero. However, unlike the scale curve, bit scales also function well when the item response probabilities approach zero. The latent trait scale resulting from any fitted IRT model can be transformed into a bit scale, aiding in score interpretation and making it easier to compare models fitted using vastly different fitting algorithms. Bit scales allows one to reap the benefits of flexible models that can adapt to the data at hand, and simultaneously obtain more interpretable latent trait scores.

The overall aim is to allow flexible IRFs while simultaneously removing any distributional assumptions on the latent trait scale, utilizing the bit scales for score interpretation. To illustrate, we use test data from the Swedish Scholastic Aptitude Test (SAT) to conduct empirical comparisons of model fit between MMC models, fitted using AEs, and NR models fitted using MML. We consider various sample sizes and test lengths throughout simulations and real data applications. Additionally, we illustrate the practical utility of bit scales in these contexts.

The remainder of this paper is structured as follows. Section \ref{sec-models} introduces the NR and MMC IRT models. Section \ref{sec-autoencoders} provides an overview of AEs and their use in fitting IRT models, followed by the introduction of bit scales in Section \ref{bit-scales}. Sections \ref{bit-empirical} and \ref{bit-simulations} present model comparisons through an empirical study and simulations. Finally, Section \ref{bit-discussion} discusses the results, practical implications, and suggestions for future studies.

\section{Item response theory models}\label{sec-models}
We assume a test with \(j=1, 2, ..., J\) multiple choice test items and \(i=1, 2, ..., N\) test takers. The response vector for test taker \(i\) is denoted 
\(\mathbf{X}_i=(X_{i1}, X_{i2}, ..., X_{iJ})\). This is a random vector where each element \(X_{ij}\) represents the chosen response option by test taker \(i\) to item \(j\). We use 
\(\mathbf{x}_i=(x_{i1}, x_{i2}, ..., x_{iJ})\) to denote its realization. In contexts where the specific test taker is redundant, we drop the subscript \(i\), thus letting 
\(\mathbf{X}=(X_{1}, X_{2}, ..., X_{J})\) be the response vector for an arbitrary test taker. Furthermore, we let \(m=1, 2, ..., M_j\) be the multiple choice options on item \(j\), possibly also including a separate option indicating a missing/invalid test taker response. We denote the IRFs for item $j$ as $p_j(m\mid\theta)=P(X_j=m\mid\theta)$ where \(\theta\) is the latent trait being measured by the item.

\subsection{IRT model assumptions}
Most IRT models are based on the following assumptions \citep{lord1980}: 1) Unidimensionality: The test should measure a single latent trait. 2) Local independence: The probability of observing a particular response for a given item should be conditionally independent of the responses to other items, given the test taker's latent trait $\theta$. Note that this assumption implicitly follows from the unidimensionality assumption, but it is often listed as a separate one. 3) Correct functional form: The IRFs should be correctly specified, reflecting the true relationships between the latent trait and the item responses. 4) Monotonicity: The probability (IRF) for the correct response should be monotonically increasing in $\theta$. 

\begin{figure}[htbp]
    \centering{
        \includegraphics[width=1\linewidth]{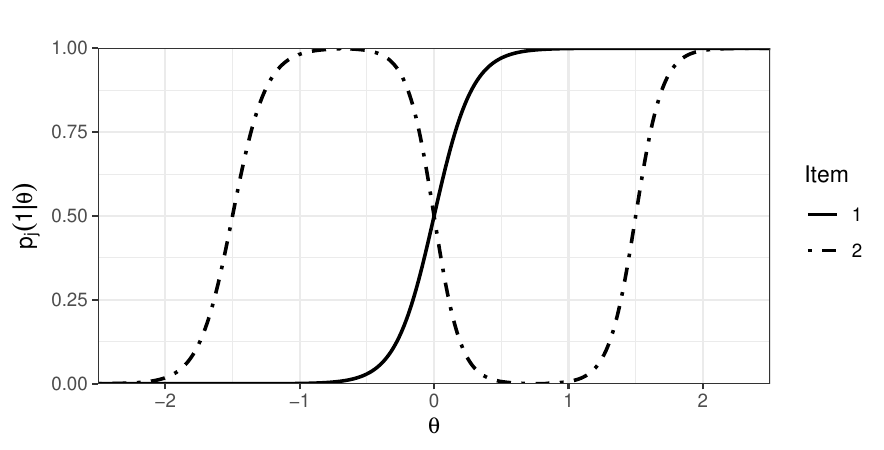}
    }
    \captionsetup{aboveskip=-8pt, belowskip=2pt}
    \caption{Example of IRFs in a situation where the monotonicity assumption is not fulfilled.}\label{fig-monotonicity}
\end{figure}

\noindent The monotonicty assumption is crucial for the type of models we consider in this study. We will put extra emphasis on this and include a simple example to explain why we think it should always be a necessity, even though IRT models have been previously proposed without it \citep{samejima1979,ramsay2020}. Consider two dichotomously scored test items measuring vastly different latent traits. Let us say one is a mathematics item and the other is a reading comprehension item. If we allow for the IRFs to vary non-monotonically, we may end up with something similar to the plot in Figure \ref{fig-monotonicity}. This model captures all four possible response patterns perfectly, but it is not useful for any practical purpose because $\theta$ has no meaning. It essentially incorporates both the mathematics and reading comprehension latent traits into a single $\theta$ scale. One could easily extend this example to tests with more items by just adding additional fluctuations in the IRFs. Non-monotonic unidimensional models have been shown in multiple studies to result in improved model fit \citep[e.g.,][]{wiberg2019,wallmark2023}, but they are in a sense "cheating" when doing so, as they are likely incorporating multiple latent traits into a unidimensional model. Therefore, we argue that the monotonicity assumption should always be enforced in IRT models. 

\label{sec-assumptions}

\subsection{Nominal response model}
The NR model \citep{bock1972} was created for test items with multiple unordered response categories and defines the IRF for choosing option \(m\) on item \(j\) as
\begin{equation}
    p_j(m\mid\theta)=
    \dfrac{\exp \left(a_{jm}\theta+b_{j m}\right)}
    {\sum_{t=1}^{M_j}\exp \left(a_{jt}\theta+b_{j t}\right)},
    \label{eq-nr}
\end{equation}
\noindent where \(a_{jm}\) and \(b_{jm}\) are item parameters tied to response option \(m\), where \(a_{jm}\) indicates how well option \(m\) discriminates between individuals with different latent trait levels.

\subsection{Monotone multiple choice model}\label{sec-mmc}
The MMC model is our proposed new model for multiple choice items. It allows for flexible non-linear relationships between the latent trait and the item responses. The MMC model is based on the NR model, Equation~\ref{eq-nr}, but replaces the linear predictors in the exponents with non-linear monotone functions of the latent trait, thus allowing for more flexible curves
\begin{equation}
    p_j(m\mid\theta)=\dfrac{
        \exp \left(z_{jm}(\theta)\right)
    }
    {
    \sum_{t=1}^{M_j}
        \exp\left(z_{jt}(\theta)\right)
    }.\label{eq-mmc}
\end{equation}
\noindent For each incorrect item response option, we define \(z_{jm}(\theta)=\tau_j\delta_{j m}(\theta)+b_{j m}\) where \(\delta_{j m}(\theta)\) is any monotone function of $\theta$. In this study, we use the sum of the outputs from a sequence of monotone neural network layers for each \(\delta_{j m}(\theta)\), as will described later in Section~\ref{monotone-layers}. To attain monotonicity, it is required that the derivative of $z_{jm}(\theta)$ with respect to $\theta$ for the correct response option is always larger than or equal to each of the individual derivatives for the incorrect options across the entire $\theta$ scale. To achieve this, we use $z_{jm}(\theta)=\tau_j\sum_{t=1}^{M_j}\delta_{jt}(\theta)+b_{jm}$ for the correct response option. Since the weight for the correct option is a sum of all the other monotone functions for that item and its own monotone function, this ensures that the curve for the correct option always increases faster than the IRFs for the other incorrect options. Simultaneously, it allows the incorrect option curves to vary freely in relation to each other. $\tau_j$ is a trained parameter that allows for some items to be negatively correlated with the latent trait. For the data used in this study, all $\tau_j$ were positive for all trained models. However, with different data, this may not always be the case.

\section{Autoencoders}\label{sec-autoencoders}
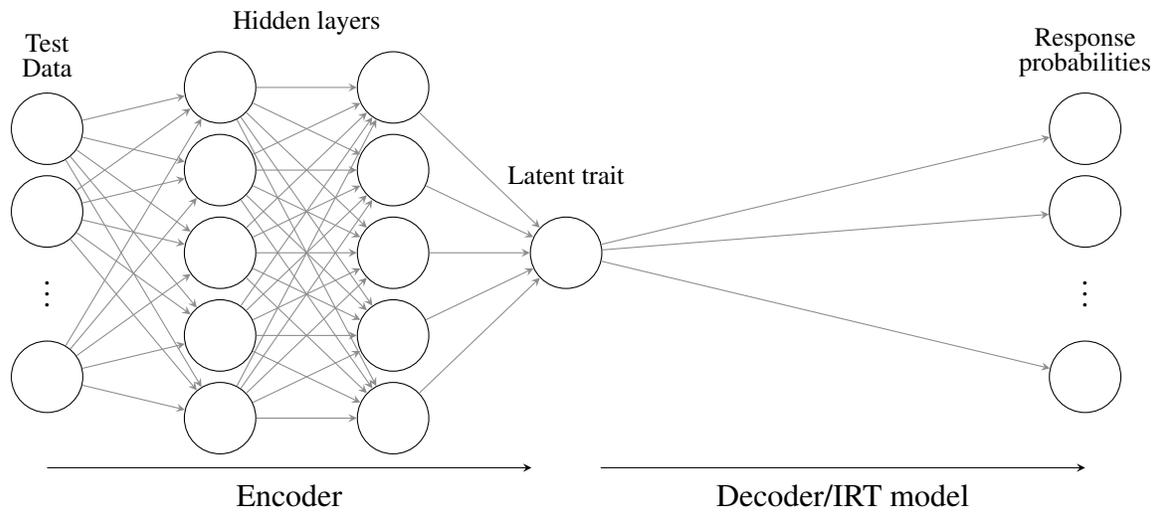
\begin{figure}[htbp]
    \centering
    \begin{tikzpicture}[x=2.3cm, y=1.1cm, >=stealth]
        \node[above, font=\large] at (3, 6) {Autoencoder};
    
        \begin{scope}[shift={(0,0)}]
            \foreach \x in {1,...,2}
            \node[circle, draw, minimum size=0.95cm] (input1\x) at (0,5-\x) {};
            \node at (0,2.1) {$\vdots$};
            \node[circle, draw, minimum size=0.95cm] (input13) at (0,1) {};
    
            \foreach \y in {1,...,5}
            \node[circle, draw, minimum size=0.95cm] (hidden11\y) at (1,\y-0.5) {};
            \foreach \z in {1,...,5}
            \node[circle, draw, minimum size=0.95cm] (hidden12\z) at (2,\z-0.5) {};
    
            \foreach \x in {1,...,2}
            \node[circle, draw, minimum size=0.95cm] (latent1\x) at (3,4-\x) {};
    
            \foreach \y in {1,...,5}
            \node[circle, draw ,minimum size=0.95cm] (hidden13\y) at (4,\y-0.5) {};
            \foreach \z in {1,...,5}
            \node[circle, draw, minimum size=0.95cm] (hidden14\z) at (5,\z-0.5) {};
    
            \node[circle, draw, minimum size=0.95cm] (output11) at (6,4) {};
            \node[circle, draw, minimum size=0.95cm] (output12) at (6,3) {};
            \node at (6,2.1) {$\vdots$};
            \node[circle, draw, minimum size=0.95cm] (output13) at (6,1) {};
    
            \draw[->] (0,  -0.1) -- (2.8, -0.1);
            \draw[->] (3.2,-0.1) -- (6,   -0.1);
    
            \node[above] at (1.4, -0.7) {Encoder};
            \node[above] at (4.6, -0.7) {Decoder};
    
            \node[above, font=\footnotesize] at (0,  4.5) {Input};
            \node[above, font=\footnotesize] at (1.5,5) {Hidden layers};
            \node[above, font=\footnotesize] at (3,  3.7) {Latent space};
            \node[above, font=\footnotesize] at (4.5,5) {Hidden layers};
            \node[above, font=\footnotesize] at (6, 4.8) {Reconstructed};
            \node[above, font=\footnotesize] at (6, 4.5) {Input};
    
            \foreach \x in {1,...,3}
            \foreach \y in {1,...,5}
            \draw[->, draw=gray!90] (input1\x) -- (hidden11\y);
            \foreach \y in {1,...,5}
            \foreach \z in {1,...,5}
            \draw[->, draw=gray!90] (hidden11\y) -- (hidden12\z);
            \foreach \y in {1,...,5}
            \foreach \z in {1,...,2}
            \draw[->, draw=gray!90] (hidden12\y) -- (latent1\z);
            \foreach \y in {1,...,2}
            \foreach \z in {1,...,5}
            \draw[->, draw=gray!90] (latent1\y) -- (hidden13\z);
            \foreach \y in {1,...,5}
            \foreach \z in {1,...,5}
            \draw[->, draw=gray!90] (hidden13\y) -- (hidden14\z);
            \foreach \z in {1,...,5}
            \foreach \x in {1,...,3}
            \draw[->, draw=gray!90] (hidden14\z) -- (output1\x);
        \end{scope}
    
        \node[above, font=\large] at (3, -2) {Autoencoder for IRT};
    
        \begin{scope}[shift={(0,-8)}]
            \foreach \x in {1,...,2}
            \node[circle, draw, minimum size=0.95cm] (input2\x) at (0,5-\x) {};
            \node at (0,2.1) {$\vdots$};
            \node[circle, draw, minimum size=0.95cm] (input23) at (0,1) {};
    
            \foreach \y in {1,...,5}
            \node[circle, draw, minimum size=0.95cm] (hidden21\y) at (1,\y-0.5) {};
            \foreach \z in {1,...,5}
            \node[circle, draw, minimum size=0.95cm] (hidden22\z) at (2,\z-0.5) {};
    
            \node[circle, draw, minimum size=0.95cm] (latent2) at (3,2.5) {};
    
            \node[circle, draw, minimum size=0.95cm] (output21) at (6,4) {};
            \node[circle, draw, minimum size=0.95cm] (output22) at (6,3) {};
            \node at (6,2.1) {$\vdots$};
            \node[circle, draw, minimum size=0.95cm] (output23) at (6,1) {};
    
            \draw[->] (0,  -0.1) -- (2.8, -0.1);
            \draw[->] (3.2,-0.1) -- (6,   -0.1);
    
            \node[above] at (1.4, -0.7) {Encoder};
            \node[above] at (4.6, -0.7) {Decoder/IRT model};
    
            \node[above, font=\footnotesize] at (0, 4.8) {Test};
            \node[above, font=\footnotesize] at (0,  4.5) {Data};
            \node[above, font=\footnotesize] at (1.5,5) {Hidden layers};
            \node[above, font=\footnotesize] at (3,  3.2) {Latent trait};
            \node[above, font=\footnotesize] at (6, 4.8) {Response};
            \node[above, font=\footnotesize] at (6, 4.5) {probabilities};
    
            \foreach \x in {1,...,3}
            \foreach \y in {1,...,5}
            \draw[->, draw=gray!90] (input2\x) -- (hidden21\y);
            \foreach \y in {1,...,5}
            \foreach \z in {1,...,5}
            \draw[->, draw=gray!90] (hidden21\y) -- (hidden22\z);
            \foreach \y in {1,...,5}
            \draw[->, draw=gray!90] (hidden22\y) -- (latent2);
            \foreach \x in {1,...,3}
            \draw[->, draw=gray!90] (latent2) -- (output2\x);
        \end{scope}
    \end{tikzpicture}
    \caption{Example of an autoencoder neural network with a 2-dimensional latent space, and an autoencoder used to fit an IRT model.}
    \label{fig-ae}
\end{figure}

To fit models with non-constrained latent trait scales, we use autoencoders \citep[AEs,][]{kramer1991}. An AE is a type of artificial neural network utilized for learning efficient encodings of input data. It consists of two main components: an encoder and a decoder, see the top plot in Figure~\ref{fig-ae}. The encoder compresses the input data and produces a condensed representation, often referred to as the latent space or bottleneck. The decoder then reconstructs the input data from this condensed representation. The goal of an AE is to minimize the difference between the original input and the reconstructed output, thereby learning a compact and useful representation of the data through the latent space. The entire model is a feed forward neural network and each part is explained below.

The encoder function of the AE takes the input data and transforms it into a lower-dimensional space. This transformation is accomplished through a series of \(L\) hidden layers, each applying a non-linear transformation \(f(\cdot)\) to its input. If we denote the input data as \(\mathbf{h}_0\) and the constricted latent representation of this data as \(\mathbf{h}_L\), the transformation at the \(l^{\text {th }}\) layer of the encoder can be represented as
\begin{equation}
    \mathbf{h}_l=f(\mathbf{W}_l \cdot \mathbf{h}_{l-1}+\mathbf{b}_l),
    \label{eq-layer}
\end{equation}

\noindent where \(\mathbf{W}_l\) and \(\mathbf{b}_l\) denote the weight matrix and bias vector at the \(l^{\text {th }}\) layer, respectively. \(f(\cdot)\) allows the model to learn non-linear relationships between the input and output data. Common choices for \(f(\cdot)\) are sigmoid, Rectified Linear Unit (ReLU) and Exponential Linear Unit (ELU) functions \citep{kramer1991, wu2020, wu2022, huang2022}. One can increase the flexibility of the encoder function by adding more hidden layers, or by increasing the number of neurons within them (more rows in \(\mathbf{W}_l\) and elements in \(\mathbf{b}_l\)). Typically, the number of neurons in the hidden layers are larger than
the number of input variables, as this allows the model to learn more complex representations of the data before reducing its dimensionality to that of the latent space.

The decoder function of an AE takes the encoded representation and transforms it back into the original input space. The procedure is the same as for the encoder, but using the latent space variables as inputs to the first layer.

\subsection{Autoencoders for IRT}\label{sec-autoencoders-irt}
AEs have been used in previous studies to fit IRT models \citep{wu2020, wu2022, paaßen2022, huang2022}. An example of an AE used to fit an IRT model is shown in the bottom plot in Figure~\ref{fig-ae}. In an IRT setting, the encoder uses the item response data as inputs and learns a latent representation of the test data, used to represent the knowledge, skills, or other latent traits of the test takers. The decoder holds all the IRFs and outputs a vector representation of the item response probabilities. Thus the AE would learn the models parameters in Equation~\ref{eq-nr}, Equation~\ref{eq-mmc}, or any other IRT model during training while simultaneously learning the weights of the hidden layers of the encoder. Note that no hidden layers would be used in the decoder unless the IRT model itself has hidden layers, such as an MMC model with the monotone layers described in Section \ref{monotone-layers}. Since the focus is on unidimensional IRT models in this study, the latent space consists of a single latent variable. One hidden layer was used for the encoder for all models fitted using AEs, as the benefits from using additional encoder layers proved close to negligible.

All AEs were trained using the \texttt{IRTorch} Python package \citep{irtorch2024}, which leverages \texttt{PyTorch} \citep{paszke2019} for parameter optimization. For the models in this study, we used stochastic gradient descent with AMSGrad \citep{reddi2019}, a variant of the ADAM optimizer, to find the AE parameters $\eta$ minimizing the negative log-likelihood loss function 

\begin{equation*} -\ell(\eta) = -\sum_{i=1}^N \log P(\mathbf{X}{i} = \mathbf{x}{i} \mid \eta). \end{equation*}

\noindent Note that $\eta$ contains the parameters of both the encoder (All \(\mathbf{W}_l\) and \(\mathbf{b}_l\) in Equation~\ref{eq-layer}) and the decoder (The item parameters in, e.g., Equation~\ref{eq-nr} or Equation~\ref{eq-mmc}). When modeling each response options separately, we are essentially doing multi-class classification with categorical distributions, which makes this loss function equivalent to the commonly used cross entropy loss function. Stochastic gradient methods iteratively update the model parameters using stochastic gradient estimates of the loss function with respect to the parameters. AMSGrad adjusts the magnitudes of parameter updates at each iteration by utilizing exponential moving averages of previous stochastic gradient estimates. To fit an AE one needs to specify two hyperparameters: (1) the batch size, which is the number of observations fed through the network in each iteration, and (2) the learning rate, the base rate of which the model parameters are updated. For a more extensive overview of stochastic gradient methods, see \citet{bottou2018}. In our experience, the models in this study tend to perform well without the need for extensive hyperparameter tuning.

When fitting IRT models using MML, one needs to integrate over an assumed latent trait distribution. Because of this, a MML fitted model is more constrained, and abilities are estimated through secondary methods like Expected A Posteriori (EAP), Maximum A Posteriori (MAP), or Maximum Likelihood (ML) \citep{Baker2004}. Any of these methods can also be used with IRT models fitted using AEs, but AE fitted models also offer a streamlined and computationally much faster alternative: Pushing the test data through the encoder function to en latent space, see Figure~\ref{fig-ae}. The encoder outputs can then be used directly as latent trait estimates without the need to estimate more parameters. We will refer to encoder latent trait estimates as neural network (NN) estimates. When the negative log-likelihood is used as the loss for fitting the AE, NN estimates are essentially Maximum Likelihood (ML) estimate approximations limited by the flexibility of the encoder network. Throughout the comparisons in this study, we will examine the relative accuracy of ML and NN methodologies in different modeling scenarios.

\subsection{Monotone layers}\label{monotone-layers}
As mentioned in Section~\ref{sec-mmc}, we used sequences of monotone neural network layers to model the monotone functions \(\delta_{j m}(\theta)\) in the MMC model. Specifically, for layer outputs $\mathbf{h}_{jm1}, \mathbf{h}_{jm2}, ..., \mathbf{h}_{jmL}$ of layers $l=1, 2, ..., L$, we have

\begin{equation*}
    \mathbf{h}_{jml}=f\left[\mathbf{W}_{jml}\cdot \mathbf{h}_{jml-1}+\mathbf{b}_{jml}\right],
\end{equation*}

\noindent with $\mathbf{h}_{jm0}=\theta$. The values of the final layer $\mathbf{h}_{jmL}$ are then summed together and used as $\delta_{jm}(\theta)$. In this setting, positive weights $\mathbf{W}_{jml}$ and a monotone activation function $f$ ensures monotonicity. To ensure positive weights and allow unconstrained optimization, we utilized the softplus function \citep{dugas2000} to re-parameterize the weights as $\mathbf{W}_{jml}=\log(1+\exp(\mathbf{W}_{jml}^*))$ and learning the $\mathbf{W}_{jml}^*$ matrices during training. Technically any monotone activation function \(f\) can be used to maintain monotonicity. Typical monotone activation functions include ReLU: \( f(y) = y \cdot \mathbf{1}_{y > 0} \) and ELU: \( f(y) = y \cdot \mathbf{1}_{y > 0} + a \left( e^{y} - 1 \right) \cdot \mathbf{1}_{y \leq 0} \), where $\mathbf{1}_{y > 0}$ and $\mathbf{1}_{y \leq 0}$ are indicator functions taking the values 1 if the condition is true and 0 otherwise. However, these functions limit the modeling to only convex relationships between items and latent traits. Instead, we used combined activation function approach as introduced by \citet{runje2023}. Specifically, we used three neurons for all hidden layers, i.e. three rows in $\mathbf{W}_{jml}^*$ and three elements in $\mathbf{b}_{jml}$. We then applied a convex function \(\breve{f}\) to the output of the first neuron, a concave function \(\hat{f}\) to the output of second neuron, and a lower/upper bounded (saturated) function \(\tilde{f}\) to the output of the third neuron, each defined as 
\begin{equation}\label{eq-act-func}
    \begin{aligned}
        & \breve{f}(y)=\text{ELU}(y) \\
        & \hat{f}(y)=-\text{ELU}(-y) \\
        & \tilde{f}(y)= \begin{cases}\breve{f}(y+1)-\breve{f}(1) & \text { if } y<0     \\
            \hat{f}(y-1)+\breve{f}(1)   & \text { otherwise. }\end{cases}
    \end{aligned}
\end{equation}

\noindent Note that more than three neurons could have been used, but we found that more neurons per layer did not result in any significant improve in model fit. The approach above gives three bias parameters $\mathbf{b}_{jml}$ and nine weight parameters $\mathbf{W}_{jml}^*$ per layer, with the exception of three weight parameters for the first hidden layer. MMC models were fitted in this fashion in both the empirical study and the simulation study. However, as previously mentioned these AE fitted IRT models assume no specific latent trait distribution. In the next section, we introduce bit scales to make the estimated latent trait scales more interpretable.

\section{Information theory and bit scales}\label{bit-scales}
As latent traits are not directly observed in the test data, IRT estimated latent trait scores can follow any distribution. Even in situations in which the latent score distributions are known a priori from modeling assumptions, distances on the score scales have no inherent meaning. In this section, we introduce what we refer to as bit scales, as a solution to this problem.

Crucial concepts for bit scales are surprisal, entropy and bits, originating from information theory \citep{shannon1948}. The transformation \(s(p)=-\log_2(p)\) converts a probability \(p\) into a quantity known as surprisal. This quantity is measured in bits and has metric properties. Surprisal is often called the self-information of an event, and due to the logarithmic transformation, it ranges from 0 to \(\infty\). If a rare event occurs, it gives us more information, leading to a higher value of surprisal. For an event that is certain to occur, with a probability of \(100\%\), the surprisal is \(-\log_2(1) = 0\). This aligns with our intuition, as we don't gain any new information from an event that was guaranteed to happen -- there is no surprise in it. The surprisal for a random event is also the number of consecutive coin flips landing heads needed to reach the probability of the event. As an illustration, an event with probability of $6.25\%$ is equivalent to the probability of 4 consecutive coin flips landing heads, as \((1/2)^4 = 6.25\%\), and thus has a surprisal of \(s(0.0625) = 4\) bits.

Entropy represents the expected surprisal of a random variable, and thus inherits the bit unit. It quantifies the average amount of information that someone would acquire from observing the outcome of a random variable. The entropy for an item \(j\) and a test taker with latent trait \(\theta\) can be computed as
\[
    H_j(\theta)=-\sum_{m=0}^{M_j} p_j(m \mid \theta)\log_2 p_j(m \mid \theta)=
    \sum_{m=0}^{M_j} p_j(m \mid \theta) s_j(m \mid \theta).
\]

As \(\theta\) changes, so does \(H_j(\theta)\). Thus, the total distance traveled in \(H_j(\theta)\) from moving from the minimum \(\theta\), denoted \(\theta^{(0)}\), to a test taker's estimated \(\theta\), \(\hat{\theta}\), can be computed as
\begin{equation}
    \begin{aligned}
        B_j(\hat{\theta})=
        \int_{\theta=\theta^{(0)}}^{\hat{\theta}}
        \left|\frac{dH_j(\theta)}{d\theta}\right| d\theta.
    \end{aligned}
    \label{eq-item-bit-score}
\end{equation}

\begin{figure}[htbp]
    \centering{
        \includegraphics[width=1\linewidth]{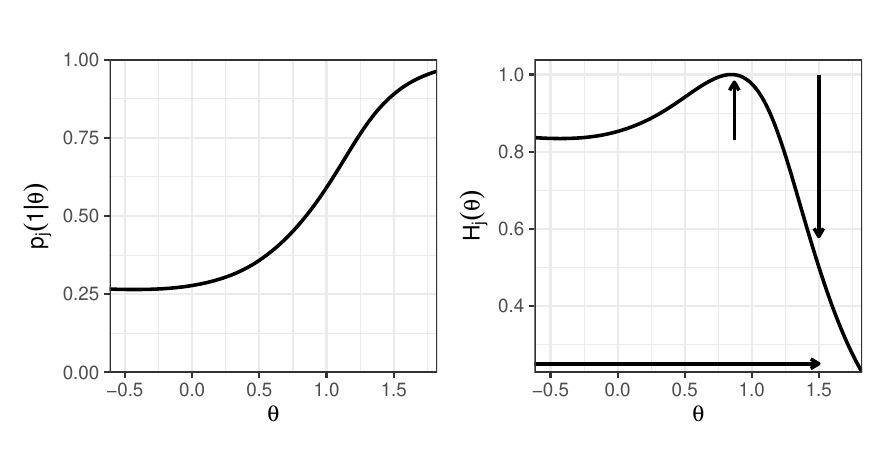}
    }
    \captionsetup{aboveskip=-8pt, belowskip=2pt}
    \caption{\label{fig-entropy-computation}IRF to the left with its
        corresponding entropy curve to the right. The vertical arrows in the
        entropy plot show the distances added up to compute the item bit
        score for a test taker with a \(\theta\) of 1.5.}
\end{figure}

\noindent These distances are what we refer to as item bit scores, located on the item bit scale. The right plot in Figure~\ref{fig-entropy-computation} illustrates how these scores are computed for a test taker with \(\hat{\theta}=1.5\) on an arbitrary dichotomously scored item. In the example, starting from the minimum \(\theta\), the entropy first increases by \(0.13\) bits, and
then decreases by \(0.49\) bits (the vertical arrows). Thus, the resulting bit score would be \(0.13+0.49=0.62\) bits, and can be interpreted as the amount of information, in bits, contained within the item a test taker at a given level of \(\theta\) has achieved. As a consequence, when \(\hat{\theta}\) equals the upper limit of the \(\theta\) scale, \(B_j(\hat{\theta})\) is the total amount of information contained within the item.

Under the local independence assumption, the item bit scores can be added together to get an information based bit score for the entire test as a function of \(\theta\), a total bit score for the test
\begin{equation}
    B(\hat{\theta})=\sum_{j=1}^J B_j(\hat{\theta}).
    \label{eq-bit-score}
\end{equation}

\noindent One should also note that both Equation~\ref{eq-item-bit-score} and Equation~\ref{eq-bit-score} are monotonic one-to-one functions of \(\theta\), and thus \(B(\hat{\theta})\) can easily be used as a \(\theta\) replacement. Furthermore, bit scales have an absolute zero, with distances measured in bits, thus resulting in a metric score scale.

\section{Empirical study}\label{bit-empirical}
The aim of the empirical study is to compare model fit between the MMC model, fitted using AEs, and the NR model, fitted using MML. AE fitted NR models were also included for additional comparisons. Real test data from the Swedish SAT administered in October 2022 were used to fit the models.

\subsection{Swedish SAT}
The Swedish SAT is offered twice a year and is used in the application process for higher education in Sweden. The test consists solely of multiple choice items and the items are publicly available after test administration\footnote{\url{https://www.studera.nu/hogskoleprov/fpn/provfragor-och-facit-hosten-2022-23-okt/}}. It includes both a quantitative and a verbal part. For this study, only the quantitative part is considered. It contains items of four different types: mathematical problem solving, quantitative comparisons, quantitative reasoning and diagrams, tables and maps. The correct response to an item is rewarded with one point and no deductions are given for replying with the wrong answer. There are 80 items in the quantitative part of the test and 38,068 test takers took the test. The mean score was $45.1$ with a standard deviation of $16.1$. There were 9,313 test takers with missing responses to one or more items.

\subsection{Model fit and evaluation measures}
MML NR models were fitted using the R package \texttt{mirt} \citep{chalmers2012}. To ensure we did not overfit the data, the dataset was randomly split into a test set and a training set. We used \(80\)\% of the data for the training set and the remaining \(20\)\% for the test set. The models were then fitted to the training set and evaluated by calculating the log-likelihood and various types of residuals using the test set.

Specifically, we use grouped residuals, comparing the observed proportions of test takers responding to each item response option within subgroups of test takers, against predictions based on the model being applied \citep{van2016b}. These subgroups were created by splitting the test takers into 10 groups based on their model estimated latent trait scores. We denote the groups $g=1, 2, ..., 10$, where the first group contains the 10\% of test takers with the lowest estimated latent trait scores, the second group contains the next 10\% of test takers, and so forth. Let $P_{gjm}$ and $p_{gjm}$ represent the observed and model predicted proportions of test takers in group $g$ responding with response option $m$ on item $j$. The grouped residual for group $g$, response option $m$ and item $j$ is then computed as $R_{gjm}=P_{gjm}-p_{gjm}$. The average of the model predicted item response option probabilities within group $g$ was used to compute $p_{gjm}$. An equal number of test takers were placed in each group.

Since we expect the magnitude of $R_{gjm}$ to vary due to sampling variability, we also computed the grouped standardized residuals $SR_{gjm}$ for each group by dividing the raw residuals with their standard deviation as follows
\begin{equation}
  SR_{gjm}= \frac{R_{gjm}}{\sqrt{p_{gjm}(1-p_{gjm})/n_{g}}},
\end{equation}
\noindent where $n_g$ is the number of test takers in group $g$. If the model fits the data well, one should expect $SR_{gjm}$ to be relatively small. 

Missing and invalid item responses were modeled together as a separate response category for each item to utilize all information in the test data. Note that this is also required to be able to incorporate missing responses when estimating the latent trait of a test taker. It does not seem plausible to ignore missing responses, as they are likely not missing at random but left blank due to not knowing the answer or running out of time. Additionally, modeling missing item responses allows one to see how their prevalence vary over the latent trait, potentially providing useful insights for test developers.

\subsection{Model selection}\label{cross-validation}

To fit IRT models through AEs, learning rate and batch size need to be chosen. 
For the MMC models, we also need to select the number of hidden layers (decoder hidden layers, not to be confused with the hidden layers in the encoder network). As stated earlier in Section~\ref{sec-autoencoders-irt}, one hidden layer was used for the encoder for all AEs. To select the remaining hyperparameters, we used 5-fold cross-validation (CV) to minimize the impact of data split randomness. five folds were used to strike a balance between bias and variance in our model evaluation, as well as to keep the computational time reasonable. For a more detailed understanding of CV, refer to \citet{hastie2009}.

A hyperparameter grid in which batch sizes \(32\), \(64\), \(128\) and \(256\) were evaluated together with learning rates \(0.02\), \(0.04\), \(0.08\) and \(0.12\). Good hyperparameters are always somewhat dependent on the data at hand, and these values were chosen based on previous experience with similar datasets and some testing on the training data. With MMC models with one, three, five and seven hidden layers were explored. Table~\ref{tbl-cv-best} shows the log-likelihood of the best performing sets of hyperparameters for each model, averaged over all five folds. The MMC model always performed better than the NR model, regardless of how many hidden layers were used. Using more than three hidden layers only resulted in a very small, if any, performance increase, together with a substantially slower fitting process. There was a larger difference between using a one layer MMC model and the NR model than between using a three-layer MMC model compared to a one-layer MMC model. As expected, there is always an improvement in fit from using ML over NN for computing the latent traits for a fitted model, although it is relatively small in these examples. The encoder network, used to retrieve the latent trait estimates with NN, does not have the flexibility to match pure ML estimation. However, it should be noted that ML requires an extra computational step and can be slow for large datasets.

\begin{table}[htbp]
  \caption{Best performing sets of hyperparameters for various IRT models fitted using AEs. Log-likelihoods were computed using ML for latent trait estimation with NN log-likelihoods shown in parentheses.}
  \label{tbl-cv-best}
  \centering
  \begin{tabular}{llllll}
    \toprule
    Model & Hidden layers & Model parameters & Learning rate & Batch size & log-likelihood \\
    \midrule
    NR    & -             & 664              & 0.04          & 256        & -80.69(-80.72) \\
    MMC   & 1             & 2404             & 0.02          & 128        & -80.09(-80.31) \\
    MMC   & 3             & 6388             & 0.04          & 64         & -79.75(-80.17) \\
    MMC   & 5             & 10372            & 0.02          & 64         & -79.78(-80.26) \\
    MMC   & 7             & 14356            & 0.02          & 128        & -79.92(-80.34) \\
    \bottomrule
  \end{tabular}
\end{table}

In general, the impact on model fit from changing the learning rate and batch size was relatively small, though learning rates of 0.08 or 0.12 for models with five and seven layers would sometimes result in failure to converge to a good solution. A batch size of 32 appeared to be a bit small for all models, resulting in worse performance. Since there were no improvements from using more than 3 hidden layers for MMC models, only AE NR, one and three-layer MMC models were fitted to the full training set using the best performing hyperparameters in Table~\ref{tbl-cv-best}.

\subsection{Empirical results}\label{empirical-results}

\begin{figure}
  \centering{
    \includegraphics[width=1\textwidth]{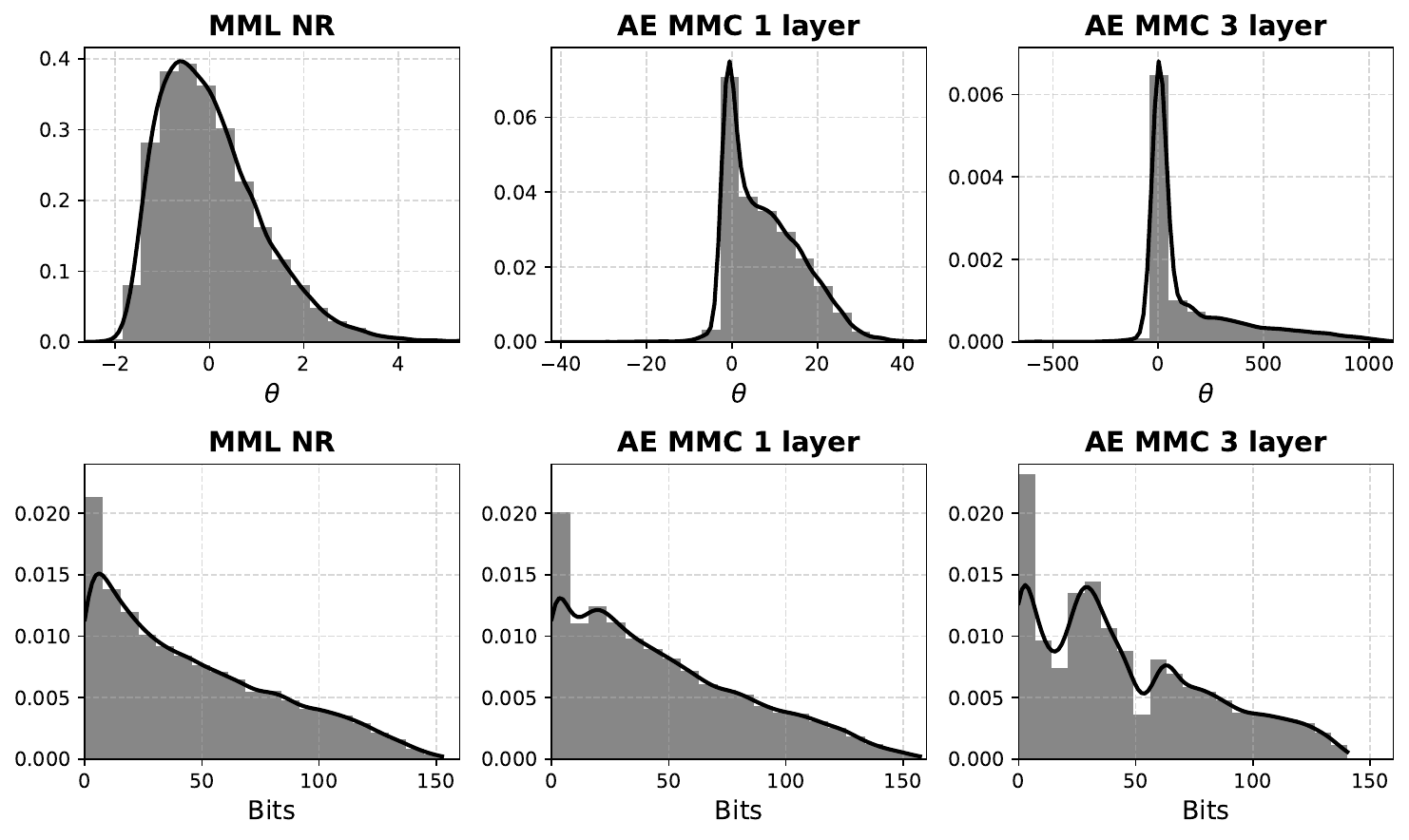}
  }
    \captionsetup{aboveskip=-8pt, belowskip=2pt}
  \caption{\label{fig-quant-distr}Histograms of \(\theta\) and bit scale
    score distributions for models fitted to the Swedish SAT.}
\end{figure}

Figure~\ref{fig-quant-distr} shows the latent trait score distributions for the test takers in the training set from various models. The starting \(\theta\) scores, \(\theta^{(0)}\), used for bit score computation were set to approximated median \(\theta\) scores of a test taker randomly guessing with equal probability for every option on all items. There is a group of test takers simply guessing on most or all items, thus rightfully receiving test scores of 0 bits.

It is clear from Figure~\ref{fig-quant-distr} that despite the normality assumption from the MML algorithm, the distribution of the MML NR model's \(\theta\) scores is a bit right skewed. The \(\theta\) scores from the MMC models have much wider score ranges compared to the NR models. For example, the top-performing test takers achieved scores exceeding 1000 from the three-layer MMC, whereas the same test takers attained scores closer to 4 when using the NR MML model. However, after being transformed to bit scores, the bit scale ranges are remarkably similar for all models despite their differences in functional form and original \(\theta\) score distributions.

The log-likelihood for each model computed on the test set is shown in Table~\ref{tbl-emp-ll}. As expected from the results in Section~\ref{cross-validation}, the MMC models fit the data better than the NR models. The NR models fitted using AEs slightly outperformed the NR models fitted using MML. These relatively small improvements indicate that choosing a more flexible IRT model is more important than allowing for less restricted latent trait scales (AE instead of MML) for this dataset. The differences between NN and ML $\theta$ estimation were found more profound for more flexible models.

\begin{table}[htbp]
  \caption{Test set log-likelihood for various IRT models computed using both ML and NN for latent trait estimation. 1hl and 3hl are for MMC models fitted using one and three hidden layers respectively.}
  \label{tbl-emp-ll}
  \centering
  \small
  \begin{tabular}{lcccc}
    \toprule
    \multicolumn{5}{c}{\textbf{Test set log-likelihood for various IRT models}} \\
    \midrule
    $\theta$ estimation & MML NR & AE NR & AE MMC 1hl & AE MMC 3hl \\
    \midrule
    ML & -80.75 & -80.68 & -80.06 & -79.75 \\
    NN & -      & -80.70 & -80.30 & -80.20 \\
    \bottomrule
  \end{tabular}
\end{table}

When comparing log-likelihood scores for separate items, the MMC models demonstrated superior fit compared to the MML NR model. The three-layer MMC model showed better fit than the MML NR model in 55 out of the 80 items, while the one-layer MMC model did so in 57. However, the three-layer MMC model still outperformed its one-layer counterpart for 41 items, as well as in overall fit. It is important to note that achieving perfect fit on one item would compromise the fit of many others if the item scores are not perfectly correlated, as all items are bound to the same latent trait scale. In a sense, the items are competing against each other in the model fitting process. In our study, this resulted in different items being harder to fit for different models. Typically, the same items would show good fit for different MMC models. However, different items might fit well with the MML NR model.

\begin{figure}
  \centering{
    \includegraphics[width=1\textwidth]{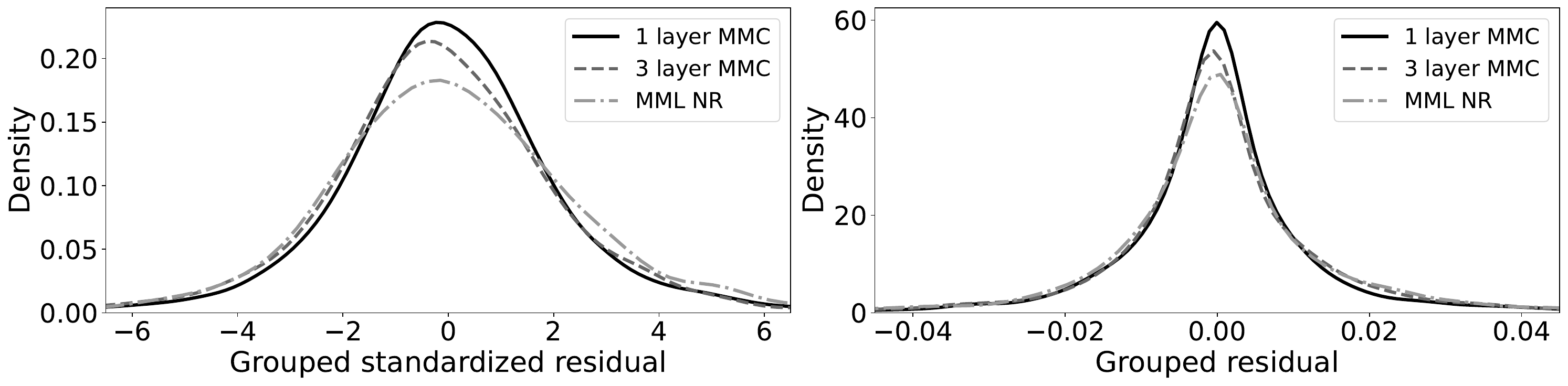}
    }
    \captionsetup{aboveskip=-8pt, belowskip=2pt}
    \caption{\label{fig-grouped-residual-hist} Kernel density estimated residual distributions for 3 different IRT models. The left plot shows the grouped standardized residuals, while the right plot shows the grouped non-standardized residuals.}
\end{figure}

Figure~\ref{fig-grouped-residual-hist} shows the distributions of the grouped and grouped standardized residuals for each model. It is evident that the MMC models provided a better fit than the NR model, as they had a larger proportion of residuals close to zero. Despite the one-layer MMC model showing worse fit  than the three-layer MMC model in terms of log-likelihood, the grouped residuals indicate that it fits the data better.

\begin{figure}[htbp]
  \centering{
    \includegraphics[width=1\textwidth]{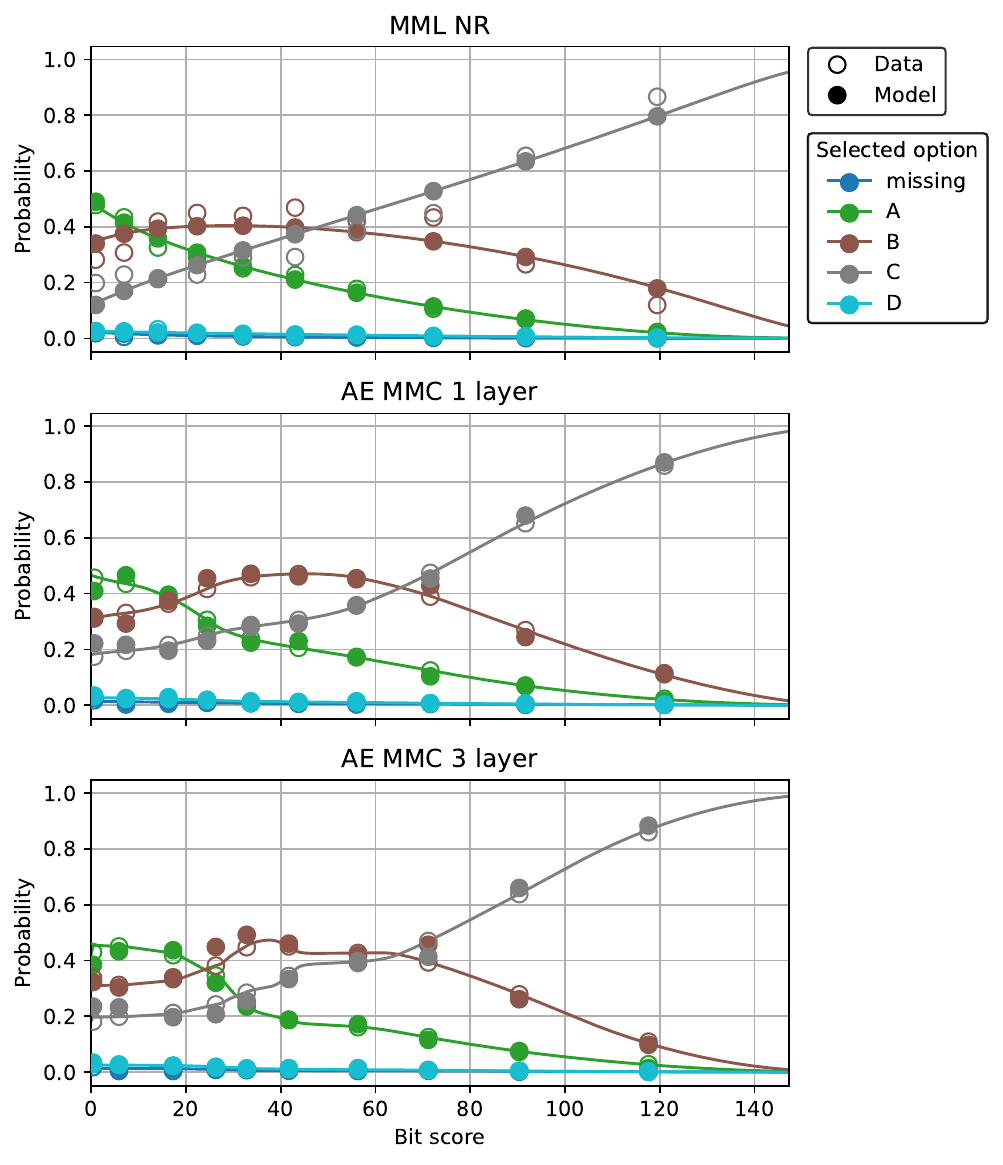}
  }
    \captionsetup{aboveskip=-8pt, belowskip=2pt}
  \caption{\label{fig-irf-nominal-bad}IRFs, along with observed proportions and model estimated proportions of test takers responding to each item response option within subgroups of test takers.}
\end{figure}

\begin{figure}[htbp]
  \centering{
    \includegraphics[width=1\textwidth]{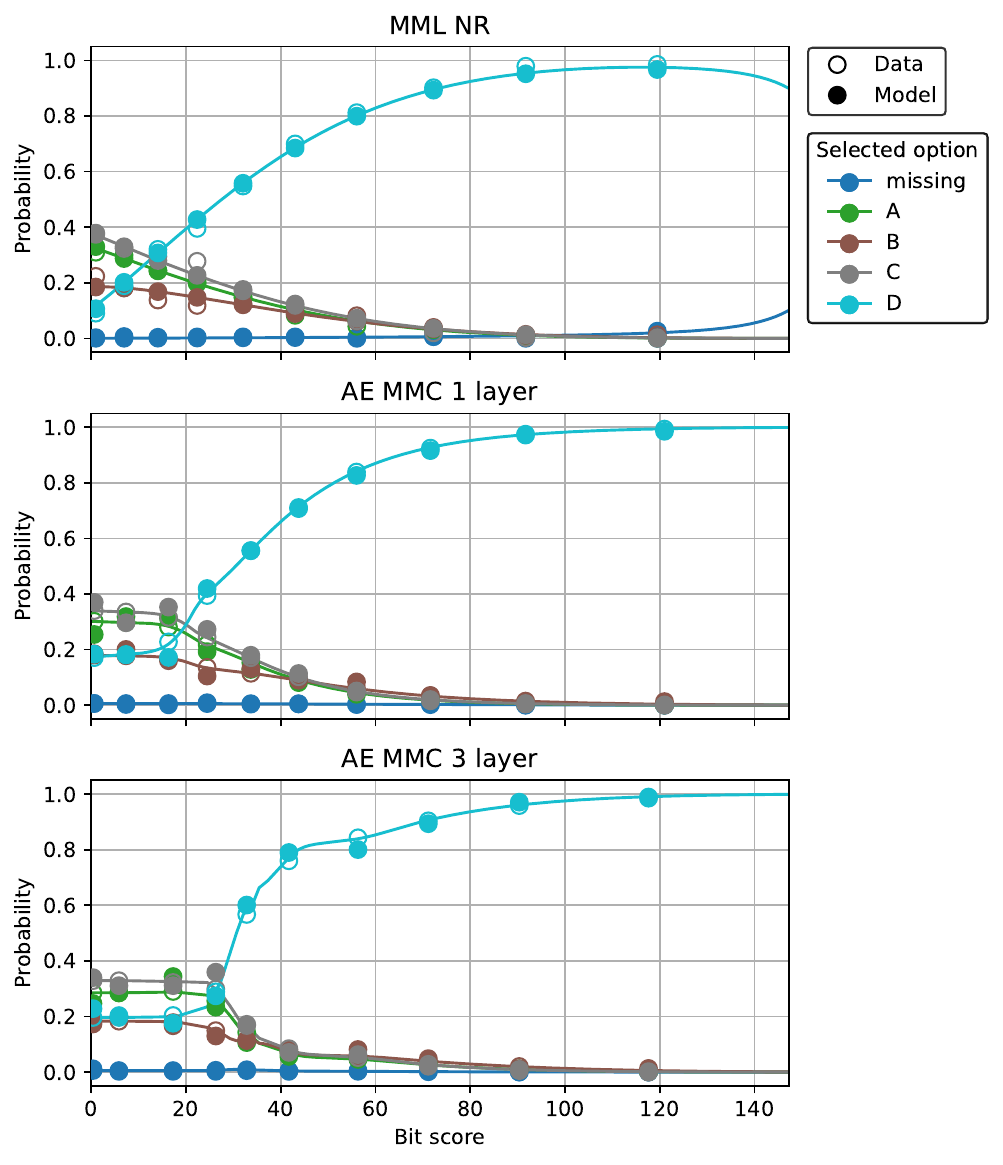}
  }
    \captionsetup{aboveskip=-8pt, belowskip=2pt}
  \caption{\label{fig-irf-nominal-failed-monotonicity}IRFs, along with observed proportions and model estimated proportions of test takers responding to each item response option within subgroups of test takers. For this item, monotonicity of the correct response option is not fullfilled for the NR model.}
\end{figure}

Figures \ref{fig-irf-nominal-bad} and \ref{fig-irf-nominal-failed-monotonicity} display comparisons of IRFs between the MML NR and the AE MMC models. The figures are plotted on the bit score scale for items that we found particularly interesting. To visualize item fit, the grouped model predicted proportions $p_{gjm}$ (the filled dots) and the observed proportions in the data $P_{gjm}$ (the see-through dots) are shown for each response option. 

\newpage
Figure~\ref{fig-irf-nominal-bad} shows an example of an item for which the MML NR model clearly did worse than the MMC models. Translated from Swedish to English, the item reads as follows:

\noindent\fbox{
  \parbox{0.97\textwidth}{
  \emph{Quantity I:} \(\frac{3^2}{7^2}\). \emph{Quantity II:} \(\frac{7^{-2}}{3^{-2}}\).
  \begin{enumerate}
    \def\labelenumi{\Alph{enumi}.}
    \setlength{\itemsep}{0pt}\setlength{\parskip}{0pt}
    \item Quantity I is greater than Quantity II
    \item Quantity II is greater than Quantity I
    \item Quantity I is equal to Quantity II
    \item the information is insufficient
  \end{enumerate}
  }
}

\noindent It appears that for this item, the lower part of the latent trait scale appears to be somewhat hard to fit for all models, but the MML NR misses the mark even more than both MMC models. Also, for this item, no matter where on the score scale a test taker is located, D has a response rate similar to that of skipping the item altogether. This is for good reason, as even if one does not know how to calculate fractions with exponents, mathematically either A, B and C has to be correct. Visualizing the IRFs in this fashion can provide useful insights, both for assessing model fit and assisting test developers in making informed decisions on whether an item or some of its response options should be rewritten.

One should note that the NR model can be somewhat flawed when analyzing
multiple choice data, as there is no guarantee that the modeled
probability to respond correctly increases with test taker ability. An
example of this is illustrated in the first plot of
Figure~\ref{fig-irf-nominal-failed-monotonicity}. It is clear that the
correct response (option D), starts to decrease as the test taker
ability increases past 120 bits (\(\theta\approx 2\)) while the
probability of a missing response increases. To suggest that test takers
with bit scores increasing beyond 120 tend to do worse on this item than
test takers with bit scores close to 120 does not seem reasonable,
especially considering that every one of the 534 test takers with
estimated bit scores above 132 in the training set got the item
correctly. Using this NR model to estimate test taker $\theta$ scores would mean  well performing test taker would be punished for getting the item correct, as the likelihood function would pull their estimated $\theta$ score towards where the probability for getting the item correct is the highest, not towards the highest possible $\theta$. Neither of the MMC models have the same issue, as it is
impossible by the functional form of the MMC IRFs. The estimated item
parameters for this NR item are \(a_{j0}=1.84\), \(a_{j1}=-1.09\),
\(a_{j2}=-0.76\), \(a_{j3}=-1.09\), \(a_{j4}=1.09\), \(b_{j0}=-3.27\),
\(b_{j1}=0.34\), \(b_{j2}=0.21\), \(b_{j3}=0.47\) and \(b_{j4}=2.25\).
seven other items also showed similar behavior. Item plots for those items
are omitted but can be obtained upon request from the corresponding author.

\section{Simulations}\label{bit-simulations}

In our simulations, we compared the model fit of MMC models fitted using AEs and NR models fitted using MML. To generate realistic multiple choice test data, observations were sampled from the quantitative part of the Swedish SAT. Sample sizes were varied between 1,000, 3,000, 5,000 and 10,000 test takers. To compare model fit for different test lengths, simulated scenarios with 20, 40 and 80 items were considered.

For all samples with 40 items, the same subset of randomly selected 40 items were used. Similarly, for all samples with 20 items, the same 20 randomly selected items were used, all being a subset of the 40 items in the 40 item scenarios. This removes the random effect from different items being chosen in different samples with the same test length.

For each of the 3 test lengths, 1,000 random samples were taken for each of the 4 sample sizes. All 3 models were fitted to each sample, and the log-likelihood of the resulting models on the non-sampled data points were computed. This ensures that we do not favour overfitted models. The simulation algorithm is summarized in Algorithm~\ref{alg:alg-simulations}.

\begin{algorithm}
\caption{Simulation Algorithm}\label{alg:alg-simulations}
\small
\begin{algorithmic}
\FOR{each test length \( j \) in \( \{20, 40, 80\} \)}
    \FOR{each sample size \( n \) in \( \{1000, 3000, 5000, 10000\} \)}
        \FOR{\( r = 1 \) to \( 1000 \)}
            \STATE Sample \( n \) test takers for test length \( j \)
            \STATE Fit NR and MMC models to the sample
            \STATE Estimate $\theta$ scores for each model using both ML and NN
            \STATE Compute log-likelihood on the non-sampled data using the model estimated probabilities
            \STATE Compute residuals (1-the probability of the observed response) on the non-sampled data
        \ENDFOR
        \STATE Average results over all \( 1000 \) samples for current \( j \) and \( n \)
        \STATE Compute the log-likelihood and residual standard errors over all \( 1000 \) samples for current \( j \) and \( n \)
    \ENDFOR
\ENDFOR
\end{algorithmic}
\end{algorithm}

In contrast to the empirical study, missing responses were not modelled in our simulations, as the NR model fitted using MML sometimes struggles to converge, depending on the sampled data. Additionally, randomly getting a sample with no missing responses on any of the items results in infinite log-likelihood for anyone with a missing response in the remaining data not used to fit the model. Furthermore, not modeling missing responses allows for comparisons complementing the empirical study, as in many situations with smaller samples modeling missing responses may not be realistic. Removing test takers with missing responses resulted in 28,755 observations to sample from out of the total 38,068.

To mimic the empirical study, we used 5-fold cross-validation to select hyperparameters for all AEs in our simulations. Only one hidden layer MMC models were considered, as the improvements from having more layers were relatively small in the empirical study, see Table~\ref{tbl-emp-ll}.

\subsection{Simulation results}
\begin{figure}[htbp]
    \centering
    \includegraphics[width=1\textwidth]{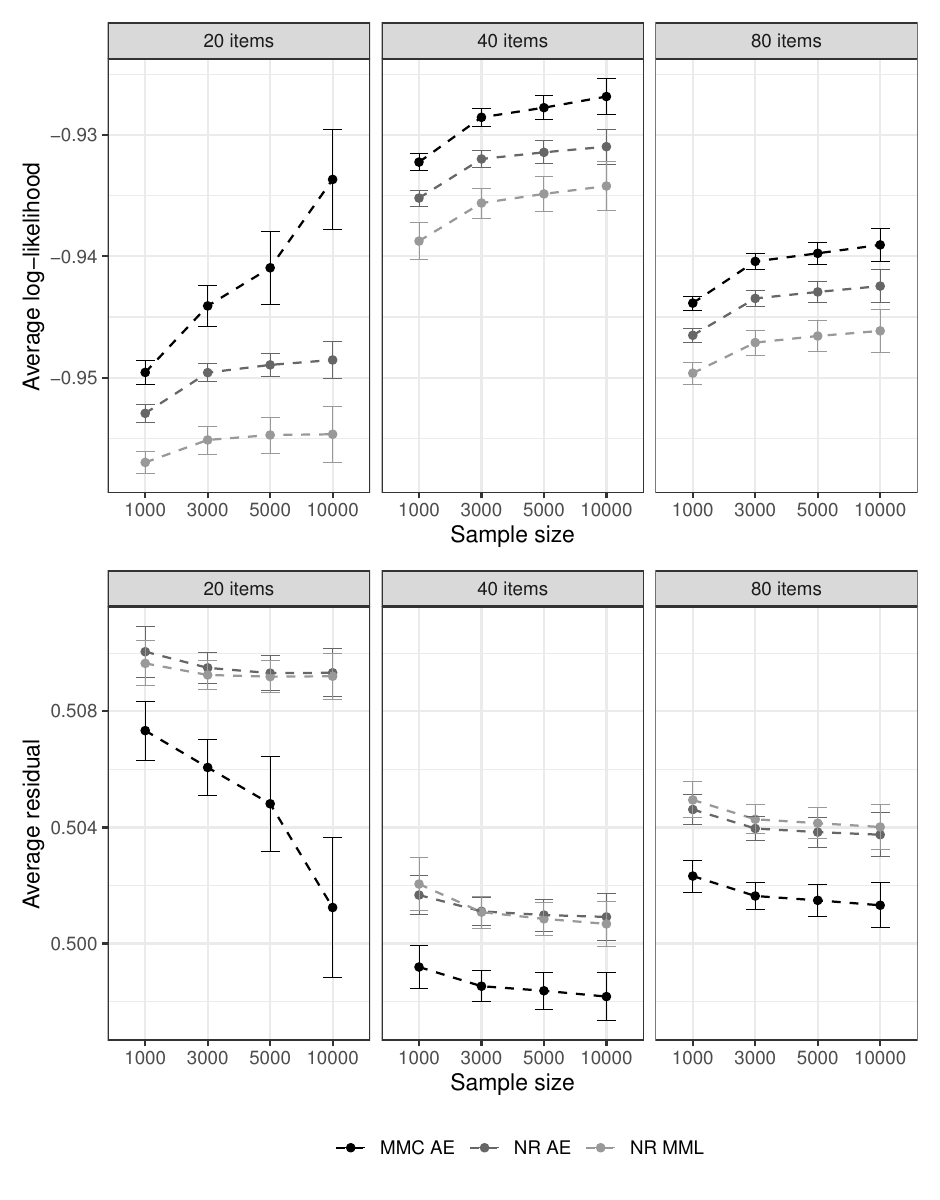}
    \captionsetup{aboveskip=-8pt, belowskip=2pt}
    \caption{\label{fig:interaction}Average test data log-likelihood and average residual in each simulated scenario. ML was used for latent trait estimation for models estimated with AEs.}
\end{figure}

The simulation results are summarized in Figure~\ref{fig:interaction} and Table~\ref{tab:simulation_results}. In all scenarios, the MMC model outperformed the NR model in terms of both residuals and log-likelihood. The MMC model showed relatively large standard errors when 20 items were used, especially for larger sample sizes. AE fitted NR models performed better than MML fitted NR models in terms of log-likelihood. This is likely due to the fact that models fitted through AEs are less restrictive about the latent trait distribution. 
\begin{table}[htbp]
    \centering
    \caption{\label{tab:simulation_results}Log-likelihood and residuals in each simulated scenario for the MMC and NR models. Standard errors are shown in parentheses.}
    \centering
    \small
    \fontsize{10}{10}\selectfont
    \begin{tabular}[t]{rrccccc}
    \toprule
    n & items & model & log-likelihood & log-likelihood & residuals & residuals\\
    & & & ML & NN & ML & NN\\
    \midrule
    1000 & 20 & MMC AE & -0.9496 (0.0010) & -0.9543 (0.0008) & 0.5073 (0.0010) & 0.5087 (0.0016)\\
    1000 & 20 & NR AE & -0.9529 (0.0007) & -0.9566 (0.0011) & 0.5100 (0.0009) & 0.5106 (0.0013)\\
    1000 & 20 & NR MML & -0.9570 (0.0009) & - & 0.5096 (0.0008) & -\\
    \midrule
    1000 & 40 & MMC AE & -0.9322 (0.0007) & -0.9352 (0.0007) & 0.4992 (0.0007) & 0.4998 (0.0013)\\
    1000 & 40 & NR AE & -0.9352 (0.0007) & -0.9379 (0.0007) & 0.5017 (0.0007) & 0.5020 (0.0012)\\
    1000 & 40 & NR MML & -0.9387 (0.0015) & - & 0.5021 (0.0009) & -\\
    \midrule
    1000 & 80 & MMC AE & -0.9439 (0.0006) & -0.9460 (0.0006) & 0.5023 (0.0006) & 0.5027 (0.0011)\\
    1000 & 80 & NR AE & -0.9465 (0.0006) & -0.9485 (0.0006) & 0.5046 (0.0005) & 0.5047 (0.0011)\\
    1000 & 80 & NR MML & -0.9496 (0.0009) & - & 0.5049 (0.0006) & -\\
    \midrule
    3000 & 20 & MMC AE & -0.9441 (0.0017) & -0.9485 (0.0010) & 0.5061 (0.0010) & 0.5076 (0.0014)\\
    3000 & 20 & NR AE & -0.9496 (0.0007) & -0.9512 (0.0008) & 0.5095 (0.0005) & 0.5100 (0.0011)\\
    3000 & 20 & NR MML & -0.9551 (0.0012) & - & 0.5092 (0.0005) & -\\
    \midrule
    3000 & 40 & MMC AE & -0.9286 (0.0007) & -0.9307 (0.0007) & 0.4985 (0.0005) & 0.4990 (0.0012)\\
    3000 & 40 & NR AE & -0.9320 (0.0007) & -0.9331 (0.0007) & 0.5011 (0.0005) & 0.5014 (0.0010)\\
    3000 & 40 & NR MML & -0.9356 (0.0012) & - & 0.5011 (0.0006) & -\\
    \midrule
    3000 & 80 & MMC AE & -0.9404 (0.0007) & -0.9418 (0.0007) & 0.5016 (0.0005) & 0.5018 (0.0011)\\
    3000 & 80 & NR AE & -0.9435 (0.0007) & -0.9445 (0.0007) & 0.5040 (0.0004) & 0.5041 (0.0010)\\
    3000 & 80 & NR MML & -0.9471 (0.0010) & - & 0.5043 (0.0005) & -\\
    \midrule
    5000 & 20 & MMC AE & -0.9409 (0.0030) & -0.9461 (0.0019) & 0.5048 (0.0016) & 0.5071 (0.0020)\\
    5000 & 20 & NR AE & -0.9489 (0.0010) & -0.9505 (0.0009) & 0.5093 (0.0006) & 0.5099 (0.0013)\\
    5000 & 20 & NR MML & -0.9547 (0.0015) & - & 0.5092 (0.0006) & -\\
    \midrule
    5000 & 40 & MMC AE & -0.9278 (0.0010) & -0.9297 (0.0010) & 0.4984 (0.0006) & 0.4990 (0.0016)\\
    5000 & 40 & NR AE & -0.9314 (0.0010) & -0.9324 (0.0009) & 0.5010 (0.0006) & 0.5014 (0.0012)\\
    5000 & 40 & NR MML & -0.9349 (0.0014) & - & 0.5009 (0.0006) & -\\
    \midrule
    5000 & 80 & MMC AE & -0.9398 (0.0009) & -0.9410 (0.0009) & 0.5015 (0.0005) & 0.5020 (0.0015)\\
    5000 & 80 & NR AE & -0.9429 (0.0009) & -0.9437 (0.0009) & 0.5038 (0.0005) & 0.5042 (0.0011)\\
    5000 & 80 & NR MML & -0.9466 (0.0013) & - & 0.5041 (0.0005) & -\\
    \midrule
    10000 & 20 & MMC AE & -0.9337 (0.0041) & -0.9390 (0.0029) & 0.5012 (0.0024) & 0.5036 (0.0025)\\
    10000 & 20 & NR AE & -0.9485 (0.0015) & -0.9496 (0.0015) & 0.5093 (0.0008) & 0.5098 (0.0014)\\
    10000 & 20 & NR MML & -0.9547 (0.0023) & - & 0.5092 (0.0008) & -\\
    \midrule
    10000 & 40 & MMC AE & -0.9268 (0.0015) & -0.9287 (0.0015) & 0.4982 (0.0008) & 0.4988 (0.0017)\\
    10000 & 40 & NR AE & -0.9310 (0.0015) & -0.9316 (0.0014) & 0.5009 (0.0008) & 0.5013 (0.0014)\\
    10000 & 40 & NR MML & -0.9342 (0.0020) & - & 0.5007 (0.0008) & -\\
    \midrule
    10000 & 80 & MMC AE & -0.9391 (0.0014) & -0.9401 (0.0013) & 0.5013 (0.0008) & 0.5017 (0.0017)\\
    10000 & 80 & NR AE & -0.9425 (0.0013) & -0.9429 (0.0013) & 0.5037 (0.0008) & 0.5041 (0.0014)\\
    10000 & 80 & NR MML & -0.9461 (0.0018) & - & 0.5040 (0.0008) & -\\
    \bottomrule
    \end{tabular}
\end{table}
However, in terms of the residuals the performance was similar. MML also appears to be a less stable fitting algorithm in these settings, as the standard errors are slightly larger for NR MML compared to NR AE in most scenarios, both in terms of log-likelihood and residuals.

One would expect that increasing the sample size would have a positive effect on model fit, and this is indeed what we observe in Figure~\ref{fig:interaction}. The effect was most profound for the MMC model in the 20 item setting. Note that the sampled items may have an effect on the results, and it appears that the 40 items sampled for the 40 item scenarios were easier to model than the 20 and 80 items. Because of this, comparing the results between scenarios differing in number of items is not meaningful.

For the AE estimated models, using the encoder neural network to estimate the latent traits (NN in Table~\ref{tab:simulation_results}) resulted in worse fit compared to using ML. However, the AE fitted models with NN latent trait estimates still outperformed the MML fitted NR models. The standard errors of the log-likelihood were similar between ML and NN, but the standard errors of the residuals were larger with NN. It is also worth noting that the MMC model outperformed the NR model fitted using MML not just on average, as shown in Table~\ref{tab:simulation_results}, but also for every single individual sample in the simulations. The simulation code can be obtained from GitHub\footnote{\url{https://github.com/joakimwallmark/mmc-bit-sim}}.

\section{Discussion}\label{bit-discussion}
In this paper, bit scales and the MMC IRT model were introduced. Through both our simulations and empirical study, we showed that using the MMC model instead of the NR model consistently leads to improved fit to the data at hand. Both in terms of log-likelihood and various types of residuals. This was true for various test lengths and number of test takers. We also showed how using AEs for model fitting outperformed MML with its implicit normality assumption, though only by a small margin. Additionally, we illustrated how bit scales can be used to provide a common metric scale that can be used not only for test takers scoring, but also for comparing scores between different models on the same scale.

We used monotone neural networks to construct the MMC model instead of other alternatives, such as the monotone splines \citep{johnson2007} or monotone polynomials \citep{falk2016}. In this case, this was a quite natural monotone function choice as the AE architecture is already a neural network. One should note that the showcased approach does result in unidentified models with no unique solution. Since the encoder network of the AE can give estimates of $\theta$ and is estimated simultaneously as the IRT model parameters, it does have some conceptual similarities with joint maximum likelihood (JML) estimation \citep{birnbaum1968}. Like JML estimation, AEs as we utilized them here are also inconsistent, as they do not converge to any particular item parameter values as the sample size increases. However, we argue that this is not a big problem, as the goal should be to approximate the true IRFs accurately as possible. Since we have no interpretation of the item parameters and the $\theta$ scale is arbitrary, model identification does not matter as much as long as the resulting model fits the data well. We consider the entire modeling procedure to be one that estimates IRFs on an arbitrarily latent trait scale; the model parameters are simply a means to that end. Even for the NR model, we would argue that plots of the IRFs, such as Figures~\ref{fig-irf-nominal-bad}-\ref{fig-irf-nominal-failed-monotonicity}, are much more informative and easier to interpret than the item parameters themselves. AEs also have an advantage over other estimation algorithms such as JML in that the number of parameters to be estimated does not increase with the number of test takers, allowing for much faster estimation for large datasets.

Adding more layers to the MMC model lead to relatively small improvements in fit compared to using a one-layer MMC instead of an NR model. We suspect that a more flexible model could, a least partially, compensate for a more constrained latent space when using fitting algorithms such as VAEs \citep{wu2022}, MML \citep[Chapter 3]{Baker2004}, or Bayesian methods \citep[Chapter 7.5]{Baker2004}. One should note that model fitting and latent trait estimation time is also substantially increased with more layers, but this may not be an issue in practice, as taking a few extra minutes to fit a model should not be a significant problem in high-stakes testing settings. Different combinations of fitting algorithms and monotone functions would likely lead to different results, and is an interesting area for future studies.

One may criticize the MMC IRFs for not capturing the lower asymptotes correctly, as it is not likely that there is always an incorrect option being chosen 100\% of the time. \citet{samejima1979} proposed a multiple choice model to solve this issue, and it would be easy to extend the MMC to also incorporate guessing probabilities in the same way. \citet{samejima1979} and \citet{thissen1984} went to great lengths to try to extrapolate the lower asymptotes of the IRFs. They provided a more complex model to achieve this, but in the process of doing so, they lost the monotoniciy assumption which we consider crucial as outlined in Section~\ref{sec-assumptions}. One should also note that a non-monotonic correct response IRF on the left penalizes examinees with low ability who respond to the item correctly. \citet{mislevy1990} introduced a modeling approach that incpororates different response strategies into the model, a potential strategy being guessing. While it would be interesting to see how this could be combined with the MMC model, we would also like to argue that scores in the lower ranges close to those who are purely guessing should hold little to no practical value. Ranking these test takers and making important decisions based on who happened to guess better seems less than ideal. In this study, we arbitrarily set the starting $\theta$ when computing bit scores ($\theta^{(0)}$ in Equation~\ref{eq-item-bit-score}) to the median estimated $\theta$ of a test taker guessing on all items with equal probability, thus effectively assigning test takers below $\theta^{(0)}$ bit scores of zero. This would essentially prevent policy makers who do not understand the models from making decisions based test takers below $\theta^{(0)}$. Of course $\theta^{(0)}$ can be set to $-\infty$ if one prefers as it makes no difference in the ordering of the scores. Nevertheless, we do not think that incorporating all the item entropy change towards the lower asymptotes when scoring test takers makes sense in the context showcased here because of the issue of one IRF always tending to 100\% probability. Although we do acknowledge that the issue of selecting $\theta^{(0)}$ would be removed if the lower asymptotes were correctly estimated.

A natural extension of this study would be to extend the MMC model for tests or questionnaires measuring multiple latent traits. Multidimensional models have already been implemented in the \texttt{IRTorch} python package. However, latent variable interpretation for multidimensional nonlinear models is challenging. For parametric models, one would typically apply some type of factor rotation method and use some of the estimated model parameters as factor loadings, see \citet{van2018}. Currently, there is no straight-forward way to do the same for a more flexible model such as the MMC, and this is also an area for future research.

The issue of the latent trait scale of an IRT model being arbitrarily defined on the scale that suits the fitting algorithm and the IRF parameters has been a long lasting problem in IRT. For example, \citet{lord1975} discussed this issue and how it relates to the item parameters, the IRF slopes and the test/item Fisher information function. The bit scale introduced in this paper provides a solution to these problems, as it is defined as an additive scale that is independent of the model parameters and modeling assumptions. The bit scale can be seen as a modification of the scale curve previously proposed by \citet{ramsay2020}. However, bit scales hold 3 key advantages over scale curves. 1) Bit scales do not tend towards infinity when the probability of responding in a given category goes towards 0. \citet{ramsay2020} solved this in the \texttt{TestGardener} R package \citep{ramsay2024} by restricting the minimum item response probabilities (maximum surpisal) in their fitting algorithm, but it makes the scale curve essentially unusable for most other models, as the surprisal values becomes infinitely large. With bit scales, this is not an issue, and it makes them more suitable to use with IRT models other than the spline models implemented in \texttt{TestGardener}. 2) A test taker's location on the bit scale is computed by adding up the item scores, while for scale curves, computation is more complex. This may appear more natural and be easier to understand for people not familiar with complicated mathematics or IRT in general. 3) With bit scales, removing or adding a test item results in constant change in the amount of information, in bits, contained within the entire test. This is more inuitive, as an item should contain the same amount of information no matter which test it is included in, as long as the underlying trait being measured is the same. Even more so when the test is given to the same test taker population. Scale curves do not have this property.

An interesting alternative for test taker scoring when the IRT models are estimated using AEs which we did not touch upon would be to use MAP or EAP \citep[Chapter 7.5]{Baker2004}. MAP and EAP allows one to incorporate information from a prior latent trait distribution, which can reduce bias for shorter tests or for test takers with extreme scores. Since there is no normality assumption with AE fitted models, one could use an NN estimated latent trait distribution as a prior. Since this distribution is learned in the fitting process, it should provide a good approximation of the true latent trait distribution. This would allow one to reap the benefits from using MAP and EAP without the need for additional constraints on the latent trait distribution.

Throughout this paper, we have introduced the MMC model and illustrated the usefulness of bit scales. In general, we believe that a better fitting model should be preferable in all cases as long as crucial assumptions are made, even if the overall differences in fit are small. Item parameters themselves hold little value and even small differences in model fit and latent trait score ordering could potentially have a profound impact on the future of an individual test taker for high-stakes tests such as the Swedish SAT. Overall, models such as the MMC model together with flexible estimation algorithms such as AEs show great promise for future applications with multiple choice type data.

\section*{Acknowledgments}
The research was funded by the Swedish Wallenberg grant MMW 2019.0129 and the Swedish Research Council grant 2022-02046.

\bibliographystyle{apalike}
\bibliography{references}

\end{document}